\documentclass[letterpaper, 10 pt, conference]{ieeeconf}  

\IEEEoverridecommandlockouts                              

\usepackage{graphicx}
\usepackage{amssymb}
\usepackage{pifont}
\usepackage{algorithm}
\usepackage{amsmath} 
\usepackage{url} 
\usepackage{hyperref}
\usepackage{algpseudocode}
\newcommand{\cmark}{\textcolor{green}{\ding{51}}}%
\newcommand{\xmark}{\textcolor{red}{\ding{55}}}%

\usepackage[table,x11names]{xcolor}

\title{\LARGE \bf
EMPOWER: Embodied Multi-role Open-vocabulary Planning with Online Grounding and Execution
}

\author{Francesco Argenziano$^{1}$, Michele Brienza$^{1}$, Vincenzo Suriani$^{2}$,  Daniele Nardi$^{1}$, and Domenico D. Bloisi$^{3}$
\thanks{$^{1}$Department of Computer, Automation and Management Engineering,
        University of Rome "La Sapienza", 00181 RM Rome, Italy
        {\tt\small lastname@diag.uniroma1.it} 
        $^{2}$School of Engineering, University of Basilicata, 85100 Potenza, Italy
        {\tt\small vincenzo.suriani@unibas.it}
        $^{3}$International University of Rome UNINT, Rome RM 00147, Italy
        {\tt\small domenico.bloisi@unint.eu}}%
}

\begin{document}

\maketitle
\thispagestyle{empty}
\pagestyle{empty}

\begin{abstract}
Task planning for robots in real-life settings presents significant challenges. These challenges stem from three primary issues: the difficulty in identifying grounded sequences of steps to achieve a goal; the lack of a standardized mapping between high-level actions and low-level commands; and the challenge of maintaining low computational overhead given the limited resources of robotic hardware.
We introduce EMPOWER, a framework designed for open-vocabulary online grounding and planning for embodied agents aimed at addressing these issues. By leveraging efficient pre-trained foundation models and a multi-role mechanism, EMPOWER demonstrates notable improvements in grounded planning and execution.
Quantitative results highlight the effectiveness of our approach, achieving an average success rate of 0.73 across six different real-life scenarios using a TIAGo robot. \end{abstract}

\section{Introduction}
A plan is a finite sequence of actions that one or more agents must execute to achieve a specific goal within a given environment. When this environment is a real-world setting, whether controlled or not, the already inherent difficulty of this task increases even more. This happens because the actions need to be \textit{grounded} in the environment, meaning a mapping is needed between the high-level actions of the plan - described either in natural language or in some formal language - and the low-level actions that the agent can execute in the physical world, in terms of joint motions. 

When those agents are real robotic systems, they typically face limitations in computational resources due to their hardware, unlike simulated agents that can run on more powerful machines. As a result, a trade-off is necessary between performance and computational overhead, which can harm the system, especially if fast execution is required. Thus, enabling an embodied agent to accurately plan and act in the world is a significant challenge in this context.
\begin{figure}
    \centering
    \includegraphics[width=1.0\columnwidth]{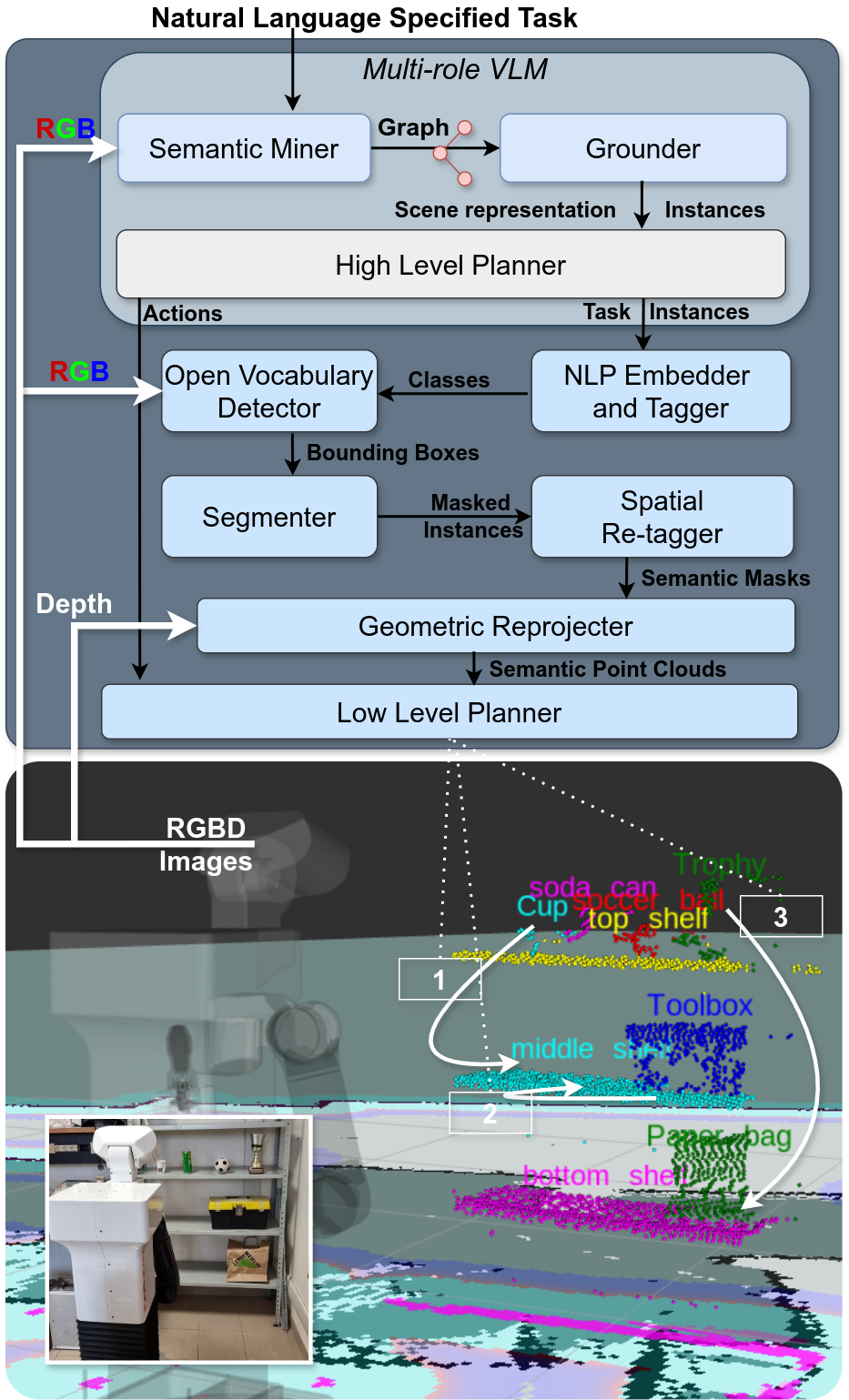}
    \caption{The EMPOWER architecture during the execution of one of the use cases analyzed: reordering the shelf to have only 2 objects per level.}
    \label{fig:schema}
\end{figure}

In recent years, we have witnessed the emergence of Foundation Models (FMs), i.e. machine learning models trained on a vast, \textit{Internet-scale} amount of data adaptable for a wide range of downstream applications \cite{bommasani2021opportunities}. Although initially designed for Natural Language Processing (NLP) tasks, these models have found utility across a variety of cross-domain applications. This happened because as a consequence of being trained on such a broad amount of data, FMs have incorporated the \textit{commonsense knowledge} which is proper to human reasoning. This knowledge, along with \textit{chain-of-thought} reasoning \cite{wei2022chain} and \textit{prompt engineering} techniques \cite{zamfirescu2023johnny}, has shown that such models, in particular, Large Language Models (LLMs) \cite{brown2020language,achiam2023gpt,touvron2023llama}, can function as few-shot planners \cite{song2023llm} and zero-shot planners \cite{huang2022language} when provided with prompts in a predefined structure. 
Recent research has shown that architectures involving multiple agents, each prompted with different roles to work together in a collaborative environment, outperform single-agent architectures \cite{talebirad2023multi, brienza2024multiagentplanningusingvisual}. 

The drawback of these architectures is that they increase the already high overhead caused by LLMs, which require substantial computational resources, such as dedicated graphics cards, not only for training but also for inference. This creates a conflict with the need for online processing.
However, recent advancements have led to the development of FMs that can operate on limited hardware within reasonable time frames, while maintaining high-quality performance, therefore enabling also their use in constrained applications. 

In this paper, we introduce EMPOWER,
a framework for task planning with robots in real-world domains (see Fig. \ref{fig:schema}). EMPOWER leverages cutting-edge FMs and cloud services to create an open-vocabulary representation of the world for efficient action grounding, and utilizes multi-role prompting techniques to generate sound plans in complex scenes.
The contributions of this work are three-fold:
\begin{enumerate}
    \item We propose a multi-role, single-body hierarchical sub-task division that enables complex reasoning about the scene and allows for the creation of plans that can be grounded in the environment;
    \item We present a complete pipeline that incorporates FMs for extracting meaningful information from the world, while being resilient to the online constraint given by the limited computational resources of real robots;
    \item We provide a flexible solution that,
    thanks to our novel open-vocabulary framework, grounds and deploys plans across different robotic platforms in the environment. 
\end{enumerate}
Quantitative results show the effectiveness of the proposed approach with a 0.73 average Success Rate (SR) across six challenging use cases performed with a TIAGo robot\footnote{https://pal-robotics.com/}.\\
The remainder of the paper is organized as follows. The subsequent Section \ref{sec:relwork} provides an overview of recent methods for robot planning. Section \ref{sec:methodology} presents the details of our approach. Qualitative and quantitative experimental results are presented in Section \ref{sec:results}. Finally, conclusions are drawn in Section \ref{sec:conclusions}. \\
Videos of the TIAGo robot's execution, code, and additional material are publicly available at \url{https://lab-rococo-sapienza.github.io/empower/}.

\section{Related Work}
\label{sec:relwork}
In this section, we review the most relevant literature on the topic. We first discuss the applications of LLMs in the field of robotics (Section \ref{sec:llmrobotics}). Next, we examine the benefits of prompt engineering, particularly in multi-role and multi-agent scenarios (Section \ref{sec:multirole}). Finally, we analyze the use of open-vocabulary detectors in complex scenarios (Section \ref{sec:openvocab}).

\subsection{LLMs and robotics}
\label{sec:llmrobotics}
Large Language Models (LLMs) \cite{brown2020language,achiam2023gpt,openai2024gpt4,touvron2023llama}, and Visual Language Models (VLMs) \cite{openai2023gpt4v, liu2023visual, geminiteam2023gemini} have emerged as some of the most significant achievements in Artificial Intelligence (AI) in recent years. Their integration with robotics has rapidly advanced, driven by the benefits these models offer. FMs have started being applied in various areas, including navigation \cite{shah2023lm}, Human-Robot Interaction \cite{zhang2023large}, conversation \cite{nwankwo2024conversation}, collaboration \cite{luan2024automatic}, and planning.

A milestone in this field is the SayCan framework \cite{ahn2022can}, which demonstrated how LLMs can serve as a planning component for embodied agents. In this framework, the robot acts as the \textit{hands and eyes} of the LLM, grounding actions to solve daily tasks in real-world environments. Subsequent works relaxed the assumption of relying solely on textual inputs, incorporating feedback from the environment either through images \cite{driess2023palm} or custom data structures \cite{rana2023sayplan}. 

Correctly prompting the LLM has also shown how LLMs can be used both as few-shot \cite{brown2020language} and zero-shot planners \cite{huang2022language}, where minimal or no training is sufficient to outperform heavily trained models and generate reliable plans for various tasks in real-world scenarios. This is largely due to the commonsense knowledge embedded in FMs. Many other works followed this research trend: Dorbala et al. \cite{dorbala2023can} demonstrated how robots can locate objects in a scene even when they are described using natural language (\textit{can you find a "Cat-Shaped" mug?}), while Yang et al. \cite{yang2023llm} illustrated how despite the \textit{bag-of-words} phenomenon \cite{yuksekgonul2022and} present in CLIP-based VLMs, an efficient vision-language-driven agent can be designed for visual navigation tasks.

Another approach involves designing specific transformer architectures \cite{vaswani2023attention} for robotics. The Robotics Transformers family \cite{brohan2022rt,brohan2023rt,padalkar2023open} has developed models with zero-shot generalization capabilities for new tasks, with ongoing efforts to create versatile robot policies that can be applied across different robots and environments.

However, combining FMs and robots presents some challenges, particularly the need for substantial computational resources\cite{bermudez2021overview}. Many robots have limited hardware, often lacking GPUs, which makes deploying LLM- or VLM-based applications difficult without external computational support. This limitation restricts the deployment of these systems to controlled setups if fast, online execution is required.

Despite these challenges, recent advances in transformer-based architectures and FMs demonstrate that it is possible to run these models autonomously on robots while maintaining state-of-the-art performance.

\begin{figure*}[t]
    \centering
    \includegraphics[width=1\linewidth]{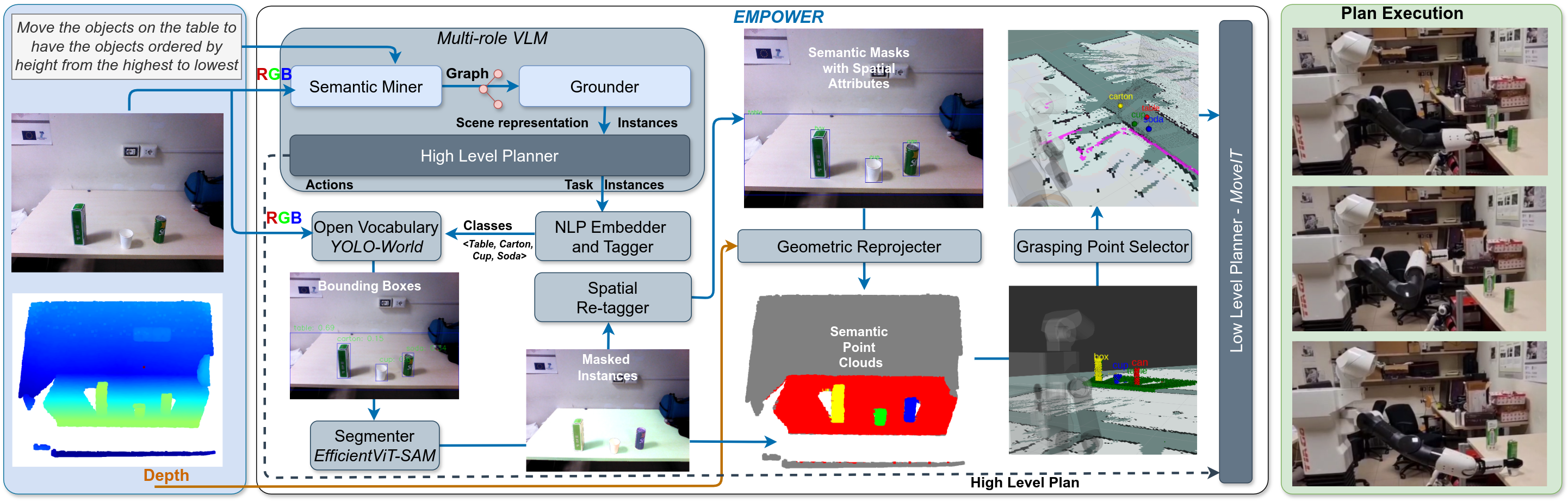}
    \caption{Complete architecture of EMPOWER, from the task description to the execution of the plan in the world. The RGB image is used to extract a graph of the scene as long as the final plan and the object labels. These labels are then grounded via an NLP pipeline and reprojected onto the point clouds extracted from the depth image of the robot. Lastly, reference points are computed from these point clouds to facilitate actions in the world. Use case illustrated: order the objects on the table from the highest to the lowest.}
    \label{fig:methodology}
\end{figure*}

\subsection{Multi-role prompting}
\label{sec:multirole}
Designing prompts to follow specific patterns has proven to be very effective, leading to better performance compared to free-form text (\textit{prompt engineering}) \cite{zamfirescu2023johnny}. Notably, Wei et al. \cite{wei2022chain} showed that asking an LLM to explicitly describe its reasoning process significantly improves the quality of the results, a technique known as \textit{chain-of-thought} reasoning. Building on this concept, other prompting strategies have been developed, such as \textit{tree-of-thoughts} \cite{yao2023tree}, which is particularly relevant in planning applications where multi-step reasoning is required. 
Despite these advancements, LLMs can still suffer from issues such as hallucinations or other undesirable outputs. A highly effective solution to this problem is leveraging multi-agent and multi-role systems. 
By following the "divide et impera" paradigm, breaking down a primary task into several sub-tasks that are then processed by multiple agents with distinct roles has produced more robust results compared to single-agent, single-role systems  \cite{wu2023autogen, rasal2024llm, shanahan2023role, talebirad2023multi, musumeci2024llm, luan2024enhancing, brienza2024multiagentplanningusingvisual}. However, designing frameworks with multiple roles and agents increases the system complexity, making them less suitable for constrained or embedded applications.
In our work, we demonstrate that the benefits gained from a multi-role architecture outweigh the additional complexity added to the system. Our system remains simple enough for fast execution on real robots in real-world settings. To the best of our knowledge, this is the first deployment of a multi-role LLM planner on a real robot.

\subsection{Open-vocabulary systems}
\label{sec:openvocab}
Recent advancements in open-vocabulary systems have contributed to bridging the semantic gap between textual descriptions and visual content, freeing models from the limitations of a fixed, closed vocabulary. Transformers have enabled open-vocabulary architectures in various tasks, including image segmentation \cite{ghiasi2022scaling, qin2023freeseg, liang2023open}, object detection \cite{zareian2021open, gu2021open, minderer2022simple} and scene parsing \cite{ding2023pla, gu2023conceptgraphs, jiang2024roboexp}. However, this level of generality comes at a cost: these models often require significant computational resources even during inference, making them less suitable for embedded applications.
In our work, we use on-the-edge models like YOLO-World \cite{cheng2024yoloworld} and EfficientViT-SAM \cite{zhang2024efficientvitsam} to achieve an open-vocabulary architecture that does not compromise the overall system's performance.

\begin{table}[t!]
\caption{Comparisons between different state-of-the-art architectures and our architecture}
\label{tab:comparison}
\begin{center}
\scalebox{1.}{
\begin{tabular}{ c | c | c | c | c | c }
\hline
\textbf{Architecture} & \textbf{Embodied} & \textbf{V.A.} & \textbf{O.V.} & \textbf{O.} & \textbf{M.R.} \\
\hline
\textbf{Huang et al.} \cite{huang2022language} & \xmark & \xmark & \xmark & - & \xmark \\
\textbf{SayCan} \cite{ahn2022can} & \cmark & \xmark & \xmark & - & \xmark \\
\textbf{LLM-Planner} \cite{song2023llm} & \xmark & \cmark & \xmark & \xmark & \xmark \\
\textbf{SayPlan} \cite{rana2023sayplan} & \cmark & \cmark & \xmark & \xmark & \xmark  \\ 
\textbf{LLM-Grounder} \cite{yang2023llm} & \xmark & \cmark & \cmark & \xmark & \xmark\\ 
\textbf{LGX} \cite{dorbala2023can} & \cmark & \cmark & \cmark & - & \xmark\\
\textbf{EMPOWER} (ours)  & \cmark & \cmark & \cmark & \cmark & \cmark\\
\hline
\end{tabular}}
\end{center}
\end{table}

Table \ref{tab:comparison} compares several state-of-the-art architectures, highlighting the presence of some key components: system embodiment in a robot (Embodied), visual assistance of the system (V.A.), open-vocabulary capabilities (O.V), online execution (O.), and the presence of a multi-role subdivision (M.R.).
To the best of our knowledge, our system is the first to integrate all of these components.

\section{Methodology}
\label{sec:methodology}
In this section, we describe the architecture of our system. Section \ref{sec:mrp} details the planner component, which utilizes a multi-role subdivision of tasks. Section \ref{sec:ovgrounding} explains the NLP processing technique we used for grounding concepts in the scene and the open-vocabulary tools we adopted. Section \ref{sec:planactuator} describes how we map the high-level grounded plan obtained in previous steps to low-level actions executable by the robot.

\subsection{Multi-role planner}
\label{sec:mrp}
The multi-role planner module consists of three agents, chosen accordingly to \cite{brienza2024multiagentplanningusingvisual}: a \emph{Semantic-Knowledge Miner Agent} (SMK) that is specialized in obtaining semantic relationships between objects in the scene; a \emph{Grounded-Knowledge Miner Agent} (GMK) that is focused on describing the environment by grounding the objects relevant to the task; a \emph{Planner Agent} (P) which is responsible for generating the final sequence of actions needed to complete the task.
For our implementation, the SMK and GMK roles are performed by GPT-4V \cite{openai2023gpt4v}, while the Planner role is handled by\cite{openai2024gpt4}. We selected these models due to their effectiveness in preliminary tests and their ability to perform efficiently without overloading the robotic system.

The agents work together to devise plans executable by a TIAGo robot. \\
Fig. \ref{fig:methodology} illustrates the interaction between these models to produce the output plan. 

\textit{\textbf{SMK}}. Takes as input an image $I$ from the robot's RGB-D camera, the description of the task that we want to achieve $d$, and it outputs a semantic representation of the scene as a set of triples: $S ={\langle\mathit{head},\mathit{relation},\mathit{tail}}\rangle$, where the \textit{relation} denotes the semantic connection between the \textit{head} and the \textit{tail} objects. This representation helps improve the robot's knowledge, similar to an instantiated Knowledge Graph of the scene \cite{hogan2021knowledge}, and assists in grounding objects perceived by the camera. 

\textit{\textbf{GMK}}. Receives the triples from \textit{SMK}, along with the image and task description. It produces a compact representation of the scene and suggests optimizations for the plan, such as renaming objects to ensure a unique representation. These unique names are used by the open-vocabulary grounder as identifiers.

\textit{\textbf{P}}. Uses the task description and the output from \textit{GMK} to generate the sequence of steps needed to accomplish the task. The planner is prompted to adhere to a predefined vocabulary of actions, facilitating the translation of the plan into low-level actions directly executable by the robot. More details on this aspect are in Section \ref{sec:planactuator}. The planning process is designed to adhere to the Markov assumption, where each action's outcome depends solely on the current state, determined through the LLM's chain-of-thought reasoning from the initial state.

\begin{table*}[]
\centering
\begin{tabular}{m{\dimexpr 0.20\linewidth-2\tabcolsep}m{\dimexpr 0.20\linewidth-2\tabcolsep}m{\dimexpr 0.20\linewidth-2\tabcolsep}m{\dimexpr 0.20\linewidth-2\tabcolsep}m{\dimexpr 0.20\linewidth-2\tabcolsep}}

\multicolumn{5}{c}{\textbf{Task Description:} Throw away the objects in the corresponding recycling bin} \\\hline
\textbf{Object World Detection} & \textbf{Spatial Tagging} & \textbf{Semantic Point cloud} & \textbf{Geometric Reprojection} & \textbf{Plan} \\
\hline
\includegraphics[width=3.5cm, height=2.5cm]{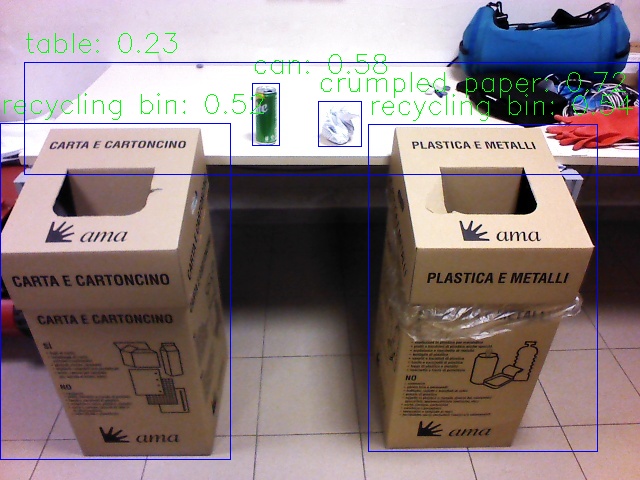} & \includegraphics[width=3.5cm, height=2.5cm]{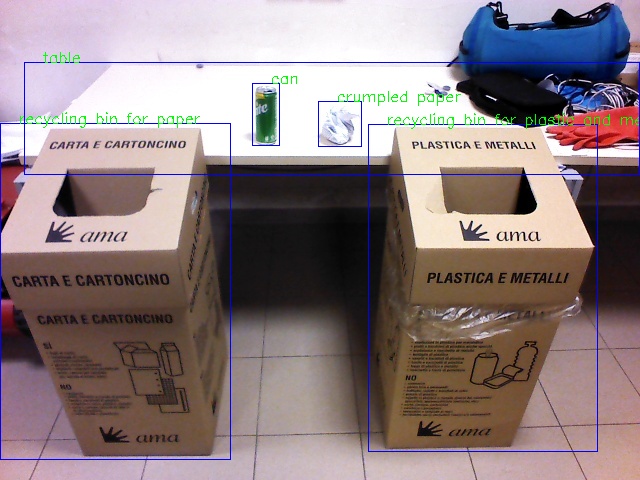} & \includegraphics[width=3.5cm, height=2.5cm]{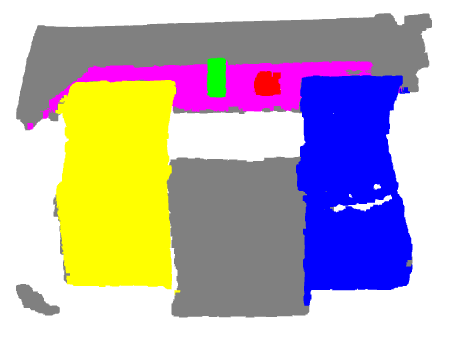} & \includegraphics[width=3.5cm, height=2.5cm]{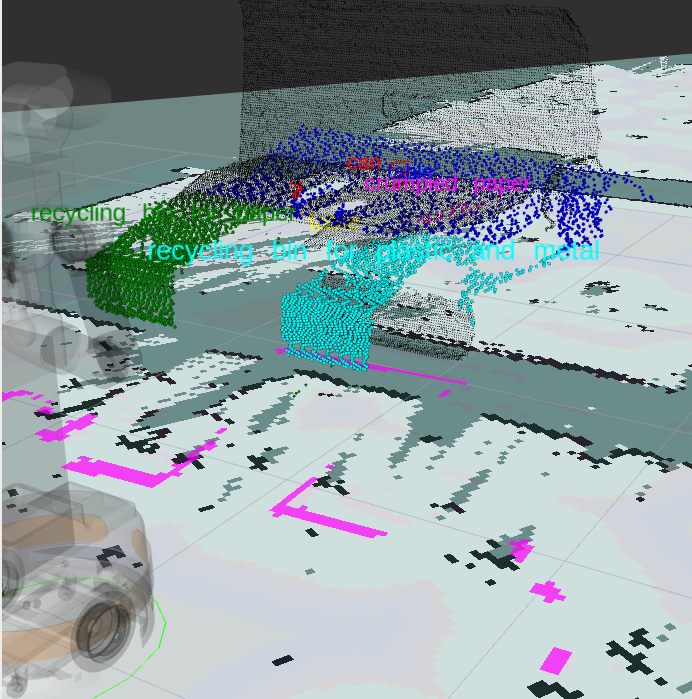} &GRAB crumpled paper, DROP crumpled paper into recycling bin for paper, GRAB can, DROP can into recycling bin for plastic and metal \\
\hline
\end{tabular}
\end{table*}

\begin{table*}[]
\centering
\begin{tabular}{m{\dimexpr 0.20\linewidth-2\tabcolsep}m{\dimexpr 0.20\linewidth-2\tabcolsep}m{\dimexpr 0.20\linewidth-2\tabcolsep}m{\dimexpr 0.20\linewidth-2\tabcolsep}m{\dimexpr 0.20\linewidth-2\tabcolsep}}

\multicolumn{5}{c}{\textbf{Task Description:} Exit the room} \\\hline
\textbf{Object World Detection} & \textbf{Spatial Tagging} & \textbf{Semantic Point cloud} & \textbf{Geometric Reprojection} & \textbf{Plan} \\
\hline
\includegraphics[width=3.5cm, height=2.5cm]{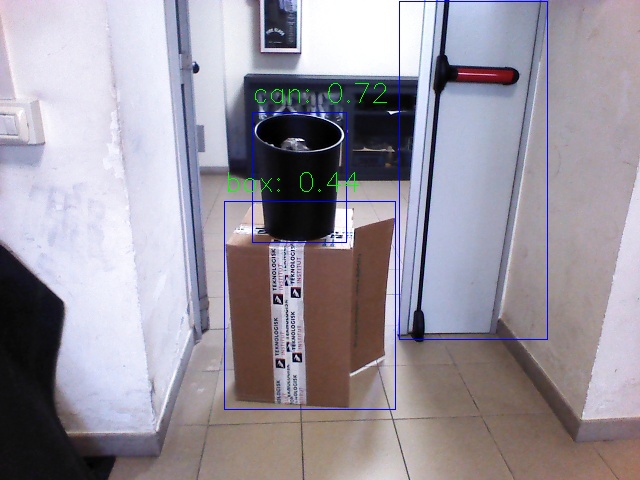} & \includegraphics[width=3.5cm, height=2.5cm]{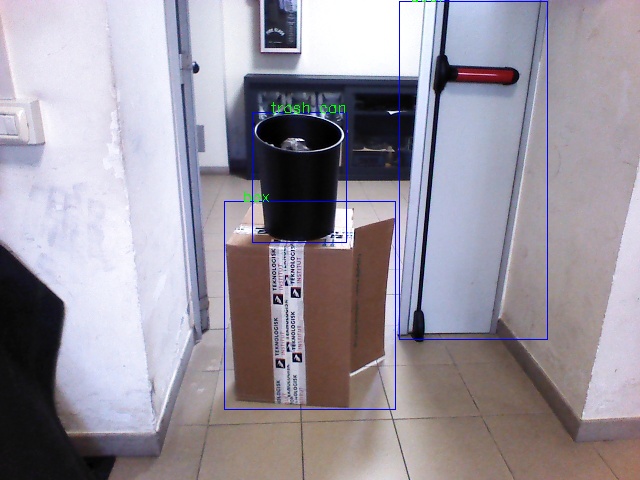} & \includegraphics[width=3.5cm, height=2.5cm]{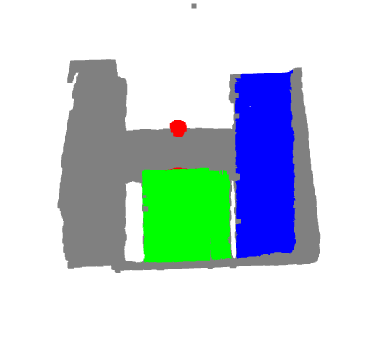} & \includegraphics[width=3.5cm, height=2.5cm]{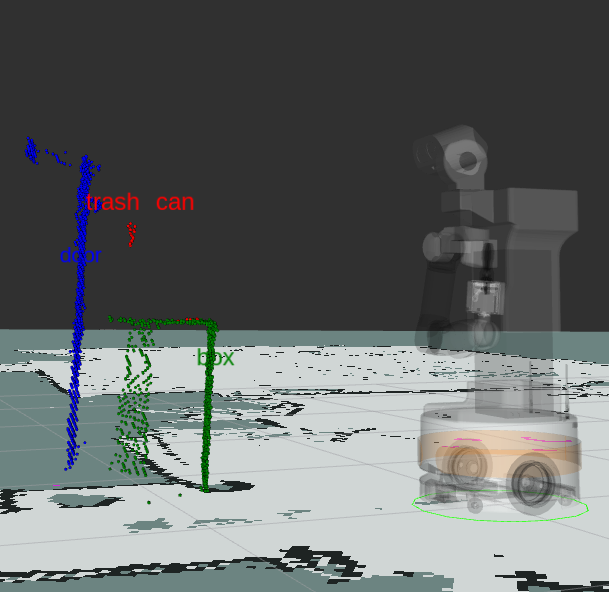} & GRAB trash can, DROP trash bin right to box, PUSH box away from the door, PULL door \\
\hline
\end{tabular}
\end{table*}

\begin{table*}[]
\centering
\begin{tabular}{m{\dimexpr 0.20\linewidth-2\tabcolsep}m{\dimexpr 0.20\linewidth-2\tabcolsep}m{\dimexpr 0.20\linewidth-2\tabcolsep}m{\dimexpr 0.20\linewidth-2\tabcolsep}m{\dimexpr 0.20\linewidth-2\tabcolsep}}

\multicolumn{5}{c}{\textbf{Task Description:} Order the shelf to have 2 objects per level} \\\hline
\textbf{Object World Detection} & \textbf{Spatial Tagging} & \textbf{Semantic Point cloud} & \textbf{Geometric Reprojection} & \textbf{Plan} \\
\hline
\includegraphics[width=3.5cm, height=2.5cm]{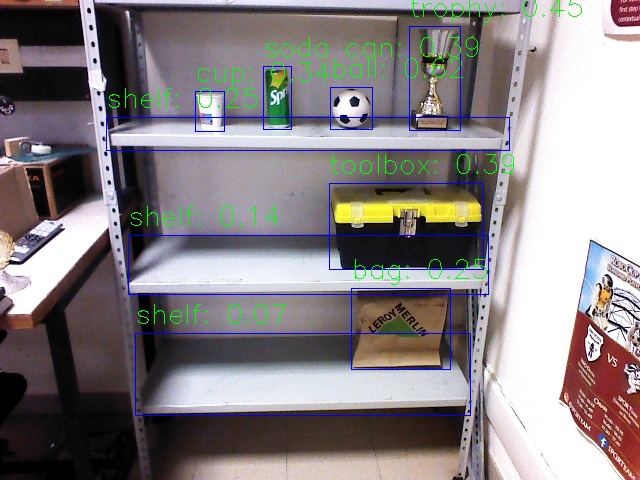} & \includegraphics[width=3.5cm, height=2.5cm]{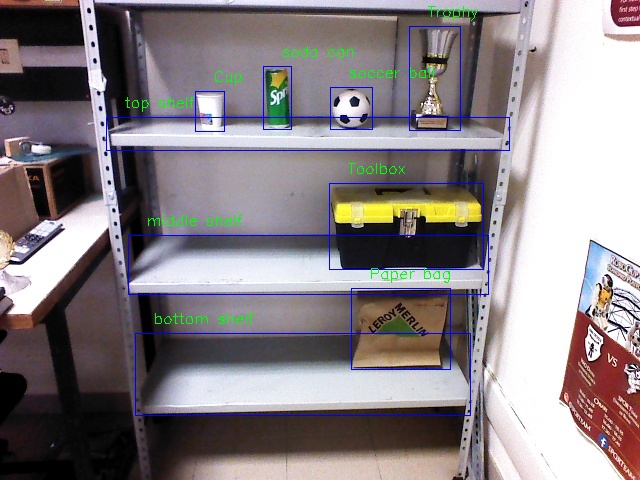} & \includegraphics[width=3.5cm, height=2.5cm]{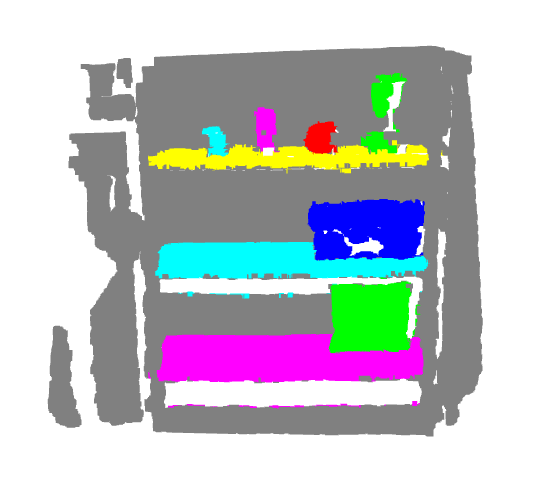} & \includegraphics[width=3.5cm, height=2.5cm]{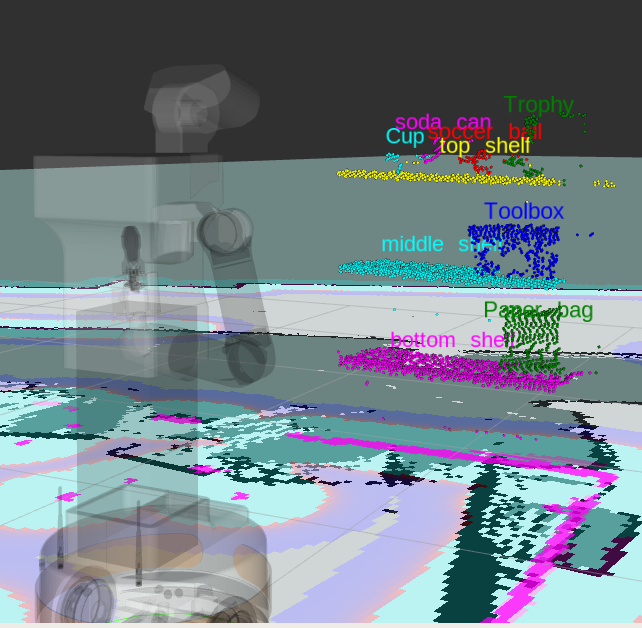} & GRAB trophy, DROP trophy middle shelf, GRAB paper cup, DROP paper cup bottom shelf \\
\hline
\end{tabular}
\end{table*}

\begin{table*}[]
\centering
\begin{tabular}{m{\dimexpr 0.20\linewidth-2\tabcolsep}m{\dimexpr 0.20\linewidth-2\tabcolsep}m{\dimexpr 0.20\linewidth-2\tabcolsep}m{\dimexpr 0.20\linewidth-2\tabcolsep}m{\dimexpr 0.20\linewidth-2\tabcolsep}}

\multicolumn{5}{c}{\textbf{Task Description:} Give me the green jacket from the clothing rack} \\\hline
\textbf{Object World Detection} & \textbf{Spatial Tagging} & \textbf{Semantic Point cloud} & \textbf{Geometric Reprojection} & \textbf{Plan} \\
\hline
\includegraphics[width=3.5cm, height=2.5cm]{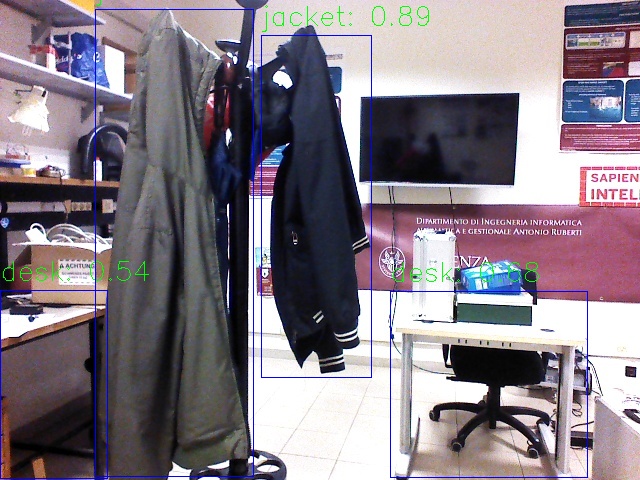} & \includegraphics[width=3.5cm, height=2.5cm]{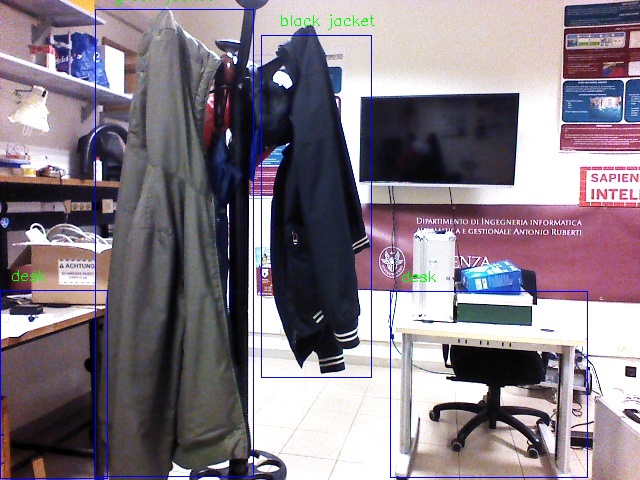} & \includegraphics[width=3.5cm, height=2.5cm]{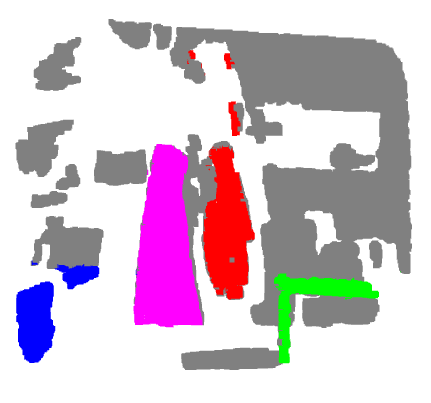} & \includegraphics[width=3.5cm, height=2.5cm]{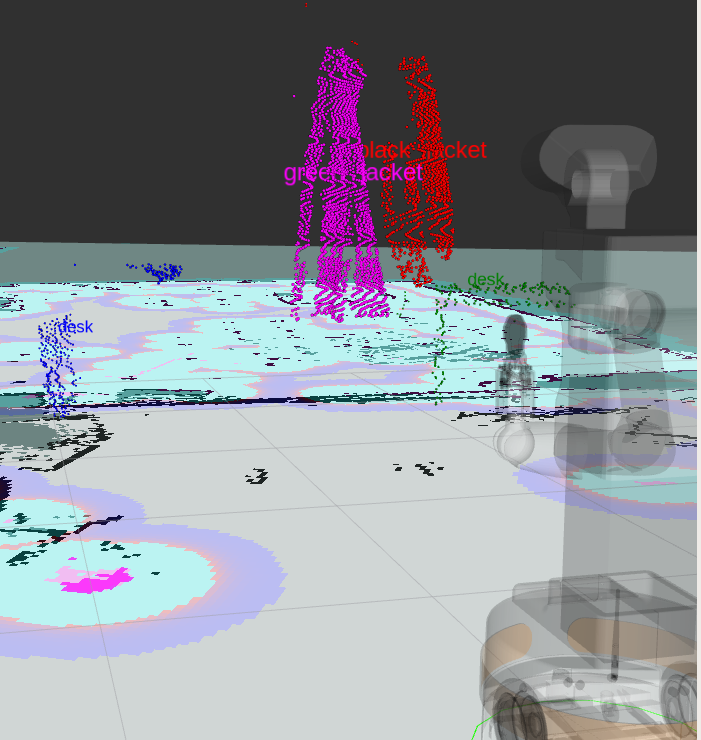} & GRAB green jacket, DROP green jacket right to you \\
\hline
\end{tabular}
\end{table*}

\begin{table*}[]
\centering
\begin{tabular}{m{\dimexpr 0.20\linewidth-2\tabcolsep}m{\dimexpr 0.20\linewidth-2\tabcolsep}m{\dimexpr 0.20\linewidth-2\tabcolsep}m{\dimexpr 0.20\linewidth-2\tabcolsep}m{\dimexpr 0.20\linewidth-2\tabcolsep}}

\multicolumn{5}{c}{\textbf{Task Description:} Move the objects on the table to have the objects ordered by height from the highest to lowest } \\\hline
\textbf{Object World Detection} & \textbf{Spatial Tagging} & \textbf{Semantic Point cloud} & \textbf{Geometric Reprojection} & \textbf{Plan} \\
\hline
\includegraphics[width=3.5cm, height=2.5cm]{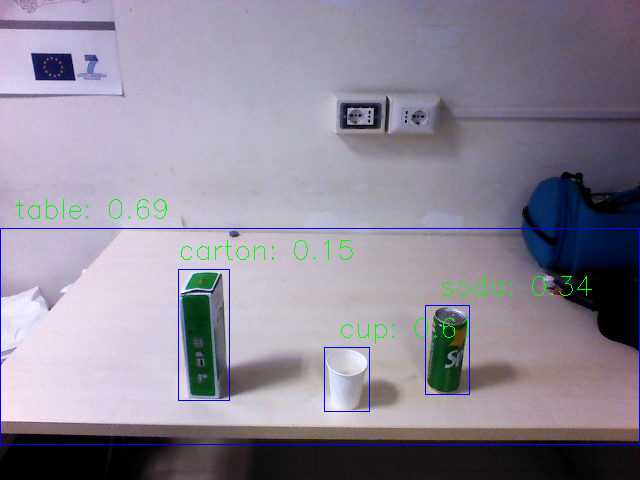} & \includegraphics[width=3.5cm, height=2.5cm]{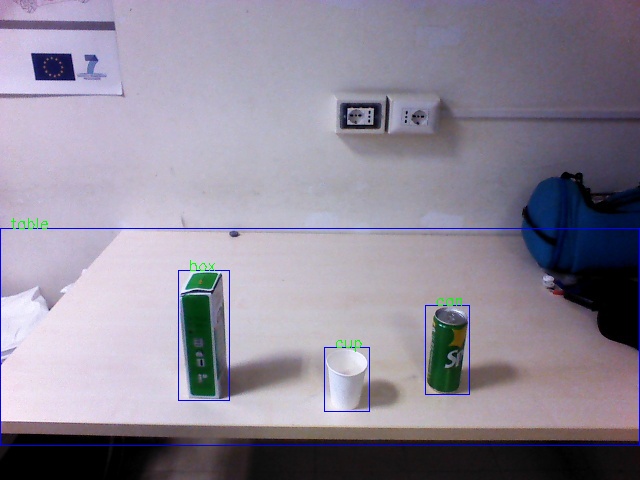} & \includegraphics[width=3.5cm, height=2.5cm]{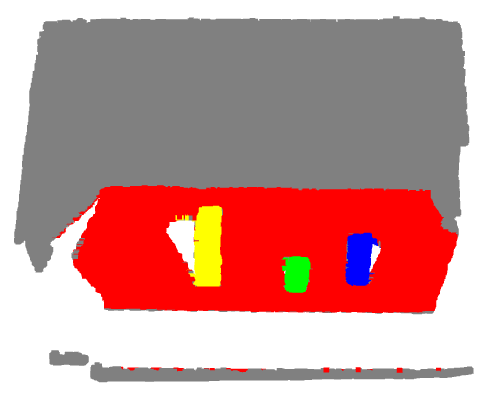} & \includegraphics[width=3.5cm, height=2.5cm]{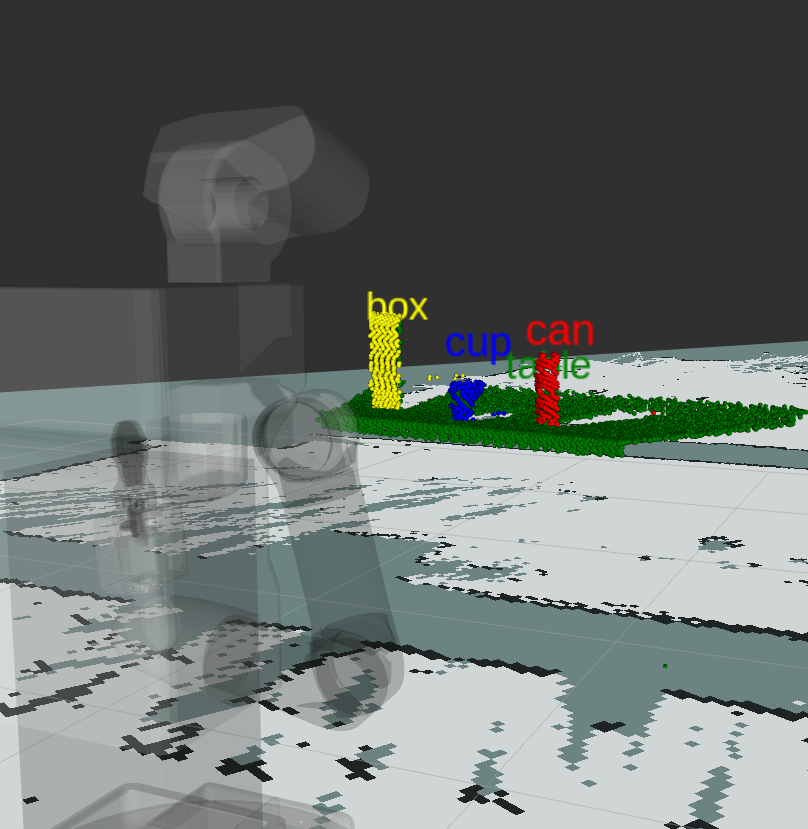} & GRAB cup, DROP cup right to can\\
\hline
\end{tabular}
\end{table*}

\begin{table*}[]
\centering
\begin{tabular}{m{\dimexpr 0.20\linewidth-2\tabcolsep}m{\dimexpr 0.20\linewidth-2\tabcolsep}m{\dimexpr 0.20\linewidth-2\tabcolsep}m{\dimexpr 0.20\linewidth-2\tabcolsep}m{\dimexpr 0.20\linewidth-2\tabcolsep}}

\multicolumn{5}{c}{\textbf{Task Description:} Move the objects on the shelf in order to have for each level of the shelf only the objects made of the same material } \\\hline
\textbf{Object World Detection} & \textbf{Spatial Tagging} & \textbf{Semantic Point cloud} & \textbf{Geometric Reprojection} & \textbf{Plan} \\
\hline
\includegraphics[width=3.5cm, height=2.5cm]{images/shelf_yolo.jpg} & \includegraphics[width=3.5cm, height=2.5cm]{images/shelf_bb.jpg} & \includegraphics[width=3.5cm, height=2.5cm]{images/shelf_mask.png} & \includegraphics[width=3.5cm, height=2.5cm]{images/shelf_geom.png} & GRAB paper cup, DROP paper cup bottom shelf, GRAB plastic ball, DROP plastic ball middle shelf, GRAB metal trophy, DROP metal trophy top shelf\\
\hline
\end{tabular}
\caption{Results of each module on all the six experimental use cases. First, the objects in the world are detected thanks to the open-vocabulary detector. Then, the spatial tagging necessary for grounding is performed. A semantic point cloud is then generated from the grounding output, and it is used to populate the RViz scene. On the right, the final plan is shown. To observe the full execution of the plans, please refer to project's website}
\label{experimental_table}
\end{table*}

\subsection{Open-vocabulary grounding}
\label{sec:ovgrounding}
The \emph{open-vocabulary grounding} component enables semantic grounding and discrimination of the objects necessary to achieve the desired task. In fact, given that multiple instances of the same object type might be present, it is crucial to distinguish their locations and roles w.r.t. the task.

The \textit{SMK} agent’s response includes a graph of objects and their spatial relations. This information helps identify instances of objects and their grounding. 

To determine object classes seen by the robot, we apply a Part-of-Speech (POS) tagging technique to extract nouns, filtering out adjectives that describe objects. We have chosen the nouns that respect the condition of central nouns, nominal subjects, or direct objects of the sentence. For instance, for an object described as a \textit{recycling bin for paper}, we only need to know that the object is a \textit{bin} to detect it: at this stage, every other information is not relevant. Therefore, the POS-tagger allows us to obtain for the object only the label that we need (\textit{bin}). We use spaCy \cite{honnibal_spacy_2018} as our POS-tagger of choice.
Objects not previously encountered are added to a list that contains all the classes needed to solve the task (Algorithm \ref{alg_objs}). We use Word2Vec similarity embeddings to check whether an object is already in the list. The use of semantic similarity is necessary because several words can describe the same object in the scene. To compute the semantic similarity between two objects, we cross-check each word of the object obtained by the POS-tagger.

\begin{algorithm}[t]
\caption{Classify Objects in Graph}
\label{alg_objs}
\begin{algorithmic}[1]
\Require $Graph = \{\langle head, relation, tail \rangle, \dots \}$
\Ensure $Classes$

\Function{alreadySeen}{$object$}
    \For{$object\_seen$ in $Classes$}
        \If{$\text{similarity}(object\_seen, object) < \tau$}
            \State $Classes.append(object)$
        \EndIf
    \EndFor
\EndFunction
\State $Classes \gets [ ]$
\For{$relation$ in $Graph$}
    \State $\text{alreadySeen}(POS\_tagger(head)$)
    \State $\text{alreadySeen}(POS\_tagger(tail)$)
\EndFor
\end{algorithmic}
\end{algorithm}

\begin{equation}
    \mathcal{E} = \frac{1}{N} \sum_{i=1}^{N} \sum_{j=1}^{M}\frac{w_i \cdot w_j}{\|w_i\| \, \|w_j\|} ,   N=|W_1| \text{ and } M=|W_2|
    \label{eq_sim}
\end{equation}
In Eq. \ref{eq_sim}, $w_i$ ad $w_j$ represent the elements in $W_1$ and $W_2$, which are the lists that contain the nouns given by POS-tagger that describe the two objects that are being compared.
For every $W_1$ and $W_2$ pair, if the similarity score $\mathcal{E}$ between two objects exceeds a threshold $\tau = 0.708$, the object is considered already detected. We have chosen this $\tau$ value using the approach presented in \cite{rekabsaz}, where they obtain thresholds that vary according to the dimensionality of the embedding vector.
This approach ensures the classes that YOLO-World can use during inference, which will then provide the bounding boxes of the objects: $\{b_i\}=\textit{YOLO-World}(I, \{l_i\})$. The automatic procedure of selecting the classes to be recognized makes our architecture open-vocabulary. These bounding boxes serve two purposes: i. locating multiple instances of the same object using spatial relations; ii. generating binary masks for each object using EfficientViT-SAM ${m_i} = \textit{EfficientViT-SAM}({b_i)}$. Lastly, these masks are reprojected onto the point cloud from the depth image to create masked point clouds for each scene object.

\subsection{Plan Actuator}
\label{sec:planactuator}
The \emph{Plan Actuator} component enables the robot to execute in the world the actions specified in the high-level plan. At this stage, we perform two key tasks: i. we \textit{refine} high-level actions by mapping them with low-level primitives that the agent can perform; ii. we \textit{ground} these actions, namely, we determine which objects the robot should interact with based on the output of the previous module. Since we are using a TIAGo robot with a gripper end-effector, we have identified a set of $5$ low-level primitives directly executable by the robot: 1. \textsc{NAVIGATE} $\langle$to a location$\rangle$, 2. \textsc{GRAB} $\langle$an object$\rangle$, 3. \textsc{DROP} $\langle$an object$\rangle$, 4. \textsc{PULL} $\langle$an object and move back$\rangle$, 5. \textsc{PUSH} $\langle$an object with the base$\rangle$. All possible action names that can appear in the plan's description are mapped to one of these $5$ primitives so that at lower levels we have more control over what action is going to be executed. 
To determine which objects the robot should interact with, we utilize the masked point clouds generated in the previous module. For each masked point cloud, we compute a representative point, the centroid. These centroids serve as the candidate locations of the grounded objects within the environment and act as the \textit{grasping points} for the robot’s actions. Finally, we use the \textit{MoveIt} \cite{coleman2014reducing} motion planning framework to deploy actions, setting the grasping point as the reference goal for the robot's end-effector.

\section{Experimental Results}
\label{sec:results}


This section details the experiments conducted in our laboratory using the robot.

\subsection{Hardware specification}
Our framework operates on the TIAGo robot, which is well-suited for indoor manipulation tasks due to its $7$ degrees-of-freedom arm. The robot is equipped with a $640$x$480$ RGB-D camera and uses a Parallel Gripper end-effector for enhanced grabbing capabilities. It is powered by $16$GB of RAM, an Intel i7 processor, $512$ GB of storage, and notably, it lacks a GPU, meaning all computations are carried out on the CPU. \\
We designed six representative use cases to evaluate the LLMs' planning capabilities for indoor tasks involving multi-step reasoning and manipulation. These use cases are: \textit{sort object by their height}, \textit{grab a jacket on the coat rack}, \textit{throw the objects in the right recycle bins}, \textit{order the shelf to have 2 objects per level}, \textit{order the shelf depending on the objects' material}, \textit{exit the room}.

\begin{figure}[t]
    \centering
    \includegraphics[width=1.0\linewidth]{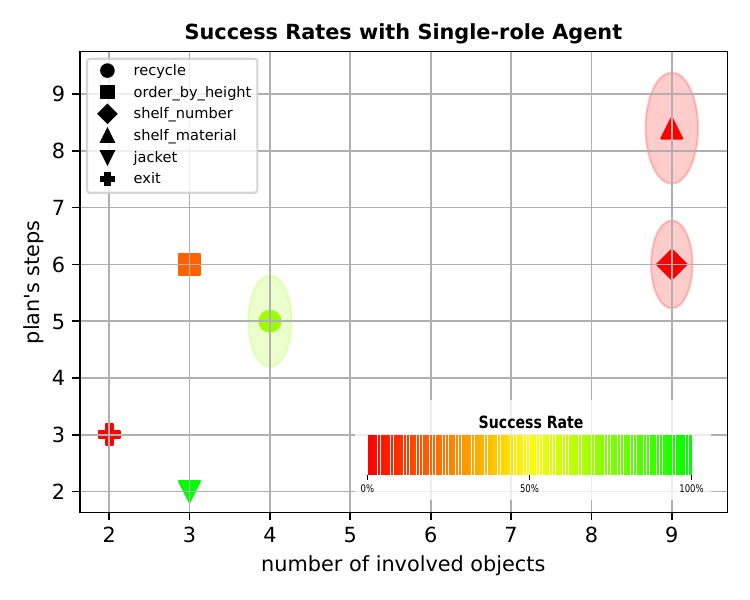}
    \caption{Success Rates of the experiments in the single-role setup. Each marker represents one particular use case. The deviation from the mean on the number of the steps of the plans over $10$ trials is also shown.}
    \label{fig:sr1}
\end{figure}

\begin{figure}[t]
    \centering
    \includegraphics[width=1.0\linewidth]{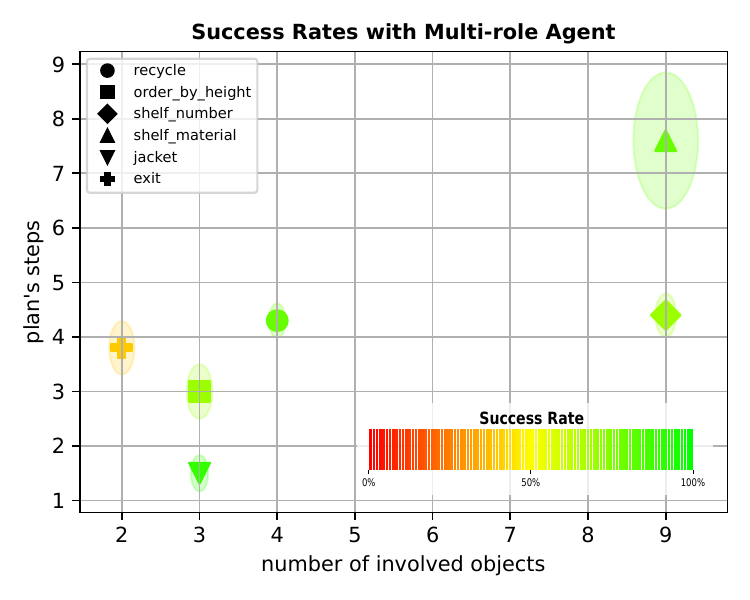}
    \caption{Success Rates of the experiments in the multi-role setup. It is possible to see how the Success Rates are higher w.r.t. the ones of the single-role setting of Fig \ref{fig:sr1}.}
    \label{fig:sr2}
\end{figure}

\subsection{Qualitative Results}
We conducted $10$ runs for each use case. Table \ref{experimental_table} shows representative results from these runs. In particular, it includes the results of the \textit{grounding} components, the semantic point cloud, their reprojections in the scene, and the plans generated by the multi-role planner. Qualitatively, the bounding boxes and binary masks demonstrate high precision in object detection. This precision enables accurate calculation of grasping points from the semantic point cloud, ensuring that the robot interacts effectively with the intended objects and executes the plans smoothly.

\subsection{Quantitative Results}
\begin{figure}[t!]
    \centering
    \includegraphics[width=1\linewidth]{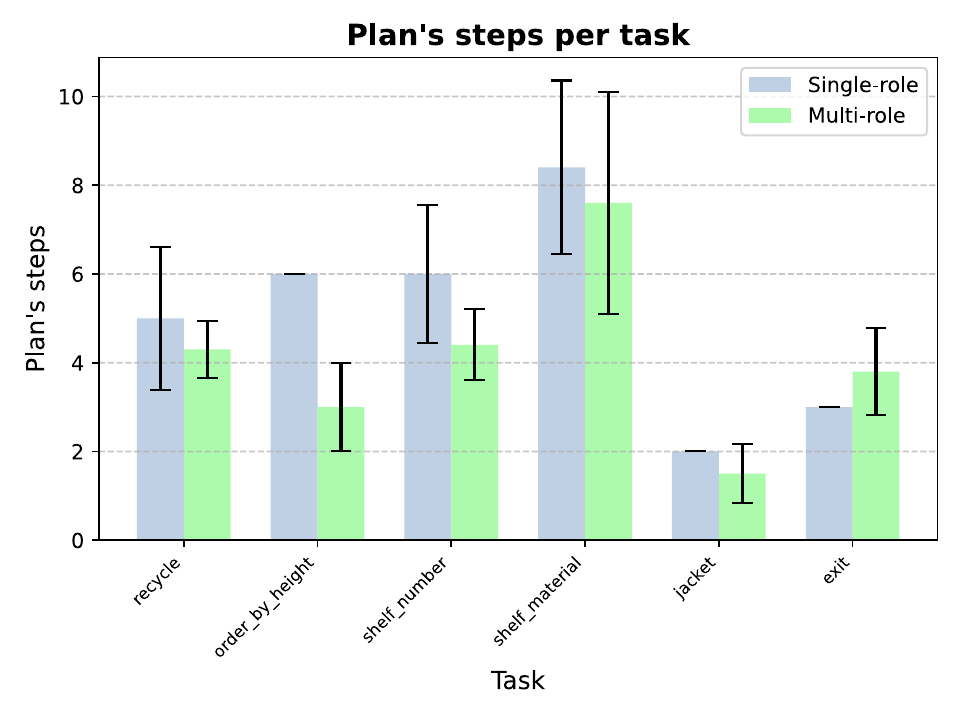}
    \caption{Comparison of average number of steps per plan between the multi-role and the single-role setup. We can see how the plans obtained by the multi-role are shorter on average than the ones of the single-role.}
    \label{fig:nsteps}
\end{figure}

To evaluate our system quantitatively, we used the Success Rate (SR) metric, which measures the percentage of successful plans across $10$ runs. A plan is deemed successful if the goal is achieved at any point and is not undone by subsequent actions; on the contrary, it is considered unsuccessful if subsequent actions nullify the achieved goal. In Figg. \ref{fig:sr1}, \ref{fig:sr2} display the results. Our comparison study shows that the multi-role architecture achieves an average SR of $73\%$, compared to $34\%$ for the single-role architecture, highlighting the effectiveness of the multi-role approach for complex reasoning tasks. An analysis of the average number of steps per plan is also shown in Fig. \ref{fig:nsteps}, and it indicates that plans generated by the multi-role setup are generally shorter and more accurate.

\subsection{Temporal Analysis}
State-of-the-art systems for open-vocabulary object detection and phrase grounding, such as GLIP \cite{li2022grounded}, require CUDA acceleration and thus cannot be used on systems without Our pipeline, however, provides comparable qualitative performance on CPU-only robots and operates efficiently on embodied systems.

We conducted a temporal analysis of our framework, examining the average execution times for the NLP pipeline, YOLO-World, and EfficientViT-SAM
The NLP pipeline takes on average in $0.8$ seconds, YOLO-World takes $0.7$ seconds, and EfficientViT-SAM processes one mask in $0.9$ seconds. Thus, the total time for a full execution ranges from $2.7$ to $8.1$ seconds, depending on the task. Therefore, the average execution time would take about $9.6$ seconds in the worst case, plus the GPT API call time, which is reasonable for real-world planning tasks.



\begin{table}[t]
\caption{Plans' Success Rates in single and multi-role architectures}
\label{succes_rates}

\begin{center}
\begin{tabular}{l | c | c}
\hline
\textbf{Task} & \textbf{Single-role} & \textbf{Multi-role}\\
\hline 
\textbf{recycle} & 0.7 & \textbf{0.8}\\
\textbf{order\_by\_height} & 0.2 & \textbf{0.7} \\
\textbf{shelf\_number} & 0.0 & \textbf{0.8} \\
\textbf{shelf\_material} & 0.0 & \textbf{0.8} \\
\textbf{jacket} & \textbf{1.0} & 0.9 \\
\textbf{exit} & 0.0 & \textbf{0.4}\\
\hline
\textbf{Average SR} & 0.32 & \textbf{0.73} \\
\hline
\end{tabular}
\end{center}
\end{table}

\section{Discussion}
This paper offers contributions both in engineering and research, advancing robot embodiment and highlighting how improved spatial and semantic awareness can enhance planning, especially in unfamiliar environments.

From an engineering perspective, the EMPOWER framework integrates pre-trained models with a multi-role planning structure, enabling real-time, open-vocabulary grounding and planning on systems like the TIAGo robot. This approach demonstrates practical, CPU-based solutions for complex tasks, making it suitable to a wide range of 
robotic platforms.

Research-wise, this work takes a first step toward more advanced robot embodiment, where robots gain a deeper semantic understanding of their environment. The multi-role architecture supports online execution, showing that enhanced awareness helps robots adapt and plan effectively, even in new settings. This shift towards open-vocabulary planning aids in better handling unexpected scenarios.

This paper underscores the value of combining spatial and semantic awareness in robotics, and we hope to pave the way for future developments in autonomous, intelligent systems.

\section{Conclusions}
\label{sec:conclusions}

In this paper, we introduced EMPOWER, a framework designed for open-vocabulary, online grounding, and planning in embodied agents, specifically aimed at addressing real-world robotic planning challenges. By leveraging efficient pre-trained foundation models within a multi-role structure, we developed a multi-role planning system that, to the best of our knowledge, is the first to support online execution in real-time settings.

Our approach enables embodied agents to plan effectively in an open-vocabulary environment, utilizing semantic understanding derived from NLP techniques applied to elements within the scene.

We evaluated our framework at multiple levels, first demonstrating the enhanced high-level planning capabilities provided by our multi-role VLM-based architecture compared to a single-role approach. We then showcased the system's overall planning capabilities, from high-level strategy to low-level execution, by testing the Success Rate of a TIAGo robot performing tasks in real environments.

The evaluation was conducted across six real-life scenarios, including tasks that require interpretative skills, which are particularly challenging for robots. For example, sorting and recycling waste by material type involves distinguishing between objects with similar appearances but different semantic meanings.
 
Quantitative results highlight the effectiveness of our approach, achieving an average Success Rate of $0.73$ across the selected challenging use cases performed by the TIAGo robot.

\section*{Acknowledgements}
This work has been carried out while Francesco Argenziano and Michele Brienza were enrolled in the Italian National Doctorate on Artificial Intelligence run by Sapienza University of Rome. \\
This work has been partially supported by PNRR MUR project PE0000013-FAIR.\\ The research reported in the paper was partially supported by the project ``Tech4You (ECS00000009) - Spoke 6", under the NRRP MUR program funded by the NextGenerationEU.









\bibliographystyle{IEEEtran}
\bibliography{references}

\end{document}